%% file: main.tex
\documentclass[runningheads]{llncs}
\usepackage{graphicx}
\usepackage{comment}
\usepackage{amsmath,amssymb} 
\usepackage{color}
\usepackage{verbatim}
\usepackage{cite}
\usepackage{fixltx2e}
\usepackage{algorithm}
\usepackage{algorithmic}
\usepackage{mathtools}
\usepackage{url}
\usepackage{hyperref}
\hypersetup{
    colorlinks=true,
    linkcolor=blue,
    filecolor=magenta,      
    urlcolor=magenta,
}
\floatname{algorithm}{Procedure}

\usepackage[width=122mm,left=12mm,paperwidth=146mm,height=193mm,top=12mm,paperheight=217mm]{geometry}

\usepackage{amsmath}
\usepackage{subcaption}
\usepackage{xcolor}
\usepackage{kotex}

\begin{document}
\pagestyle{headings}
\mainmatter
\def\ECCVSubNumber{5041}  

\title{Probabilistic Anchor Assignment with IoU Prediction for Object Detection} 

\titlerunning{Probabilistic Anchor Assignment with IoU Prediction for Object Detection}
%
\author{Kang Kim\inst{1}\thanks{Work done while at Qualcomm Korea YH.} \and
Hee Seok Lee\inst{2}}
\authorrunning{K. Kim and H.S. Lee.}
%
\institute{XL8 Inc.\\
\email{kai@xl8.ai}\and
Qualcomm Korea YH\\
\email{heeseokl@qti.qualcomm.com}}
\maketitle
\begin{abstract}
In object detection, determining which anchors to assign as positive or negative samples, known as \textit{anchor assignment}, has been revealed as a core procedure that can significantly affect a model's performance. In this paper we propose a novel anchor assignment strategy that adaptively separates anchors into positive and negative samples for a ground truth bounding box according to the model's learning status such that it is able to reason about the separation in a probabilistic manner. To do so we first calculate the scores of anchors conditioned on the model and fit a probability distribution to these scores. The model is then trained with anchors separated into positive and negative samples according to their probabilities. Moreover, we investigate the gap between the training and testing objectives and propose to predict the Intersection-over-Unions of detected boxes as a measure of localization quality to reduce the discrepancy.
The combined score of classification and localization qualities serving as a box selection metric in non-maximum suppression well aligns with the proposed anchor assignment strategy and leads significant performance improvements. The proposed methods only add a single convolutional layer to RetinaNet baseline and does not require multiple anchors per location, so are efficient. Experimental results verify the effectiveness of the proposed methods. Especially, our models set new records for single-stage detectors on MS COCO test-dev dataset with various backbones. Code is available at \url{https://github.com/kkhoot/PAA}.

\end{abstract}

\section{Introduction}
Object detection in which objects in a given image are classified and localized, is considered as one of the fundamental problems in Computer Vision. Since the seminal work of R-CNN\cite{rcnn}, recent advances in object detection have shown rapid improvements with many innovative architectural designs \cite{fasterrcnn, fpn, refinedet, m2det}, training objectives \cite{fastrcnn, focal, aploss, giou} and post-processing schemes \cite{softnms, learningnms, iounet} with strong CNN backbones\cite{lecun, alexnet, vgg, googlenet, resnet, resnext, deformableconv}. For most of CNN-based detectors, a dominant paradigm of representing objects of various sizes and shapes is to enumerate anchor boxes of multiple scales and aspect ratios at every spatial location. In this paradigm, \textit{anchor assignment} procedure in which anchors are assigned as positive or negative samples needs to be performed. The most common strategy to determine positive samples is to use Intersection-over-Union (IoU) between an anchor and a ground truth (GT) bounding box. For each GT box, one or more anchors are assigned as positive samples if its IoU with the GT box exceeds a certain threshold. Target values for both classification and localization (i.e. regression offsets) of these anchors are determined by the object category and the spatial coordinate of the GT box.

Although the simplicity and intuitiveness of this heuristic make it a popular choice, it has a clear limitation in that it ignores the actual \textit{content} of the intersecting region, which may contain noisy background, nearby objects or few meaningful parts of the target object to be detected. Several recent studies\cite{freeanchor, guidedanchor, atss, mal, noisy} have identified this limitation and suggested various new anchor assignment strategies. These works include  selecting positive samples based on the detection-specific likelihood\cite{freeanchor}, the statistics of anchor IoUs\cite{atss} or the cleanness score of anchors\cite{mal, noisy}. All these methods show improvements compared to the baseline, and verify the importance of anchor assignment in object detection.

\begin{figure}[t]
	\begin{center}
		\includegraphics[width=1.0\linewidth]{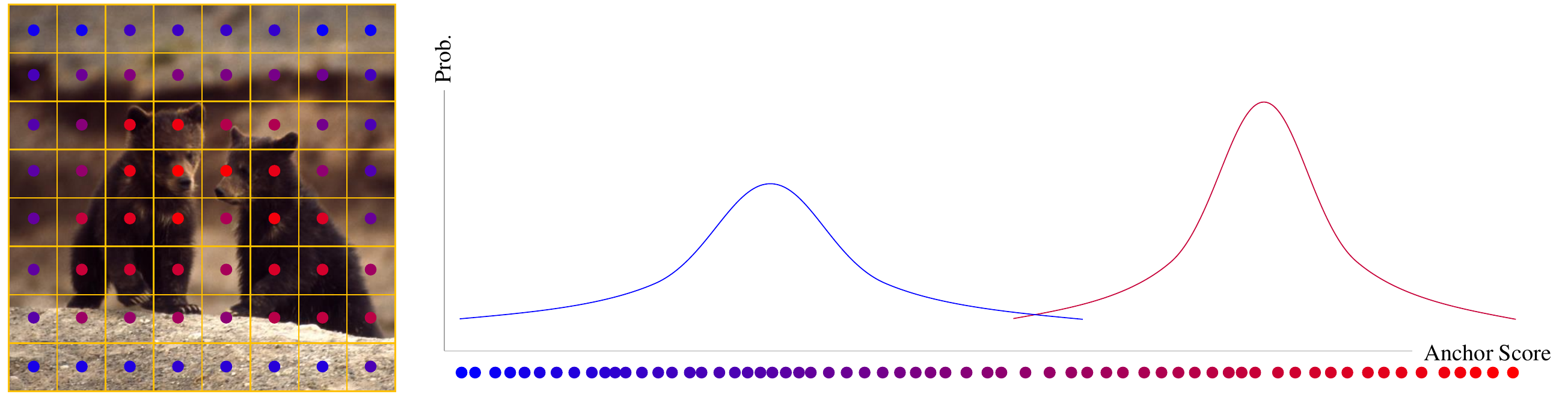}
	\end{center}
	\caption{An examplary case of anchor scores calculated by a detector model and their distribution. The scores are based on the loss objectives of classification and localization to reflect how each anchor contains meaningful cues identifiable by the model to detect a target object. We model the scores as samples from a probability distribution using Gaussian Mixture Model of two modalities (one for positive and the other for negative samples). Anchors are assigned as positive or negative samples according to their probabilities. Image source: \cite{bears}}
	\label{fig1}
\end{figure}

In this paper we would like to extend some of these ideas further and propose a novel anchor assignment strategy. In order for an anchor assignment strategy to be effective, a flexible number of anchors should be assigned as positives (or negatives) not only on IoUs between anchors and a GT box but also on how probable it is that a model can reason about the assignment. In this respect, the model needs to take part in the assignment procedure, and positive samples need to vary depending on the model. When no anchor has a high IoU for a GT box, some of the anchors need to be assigned as positive samples to reduce the impact of the improper anchor design. In this case, anchors in which the model finds the most meaningful cues about the target object (that may not necessarily be anchors of the highest IoU) can be assigned as positives. On the other side, when there are many anchors that the model finds equally of high quality and competitive, all of these anchors need to be treated as positives not to confuse the training process. Most importantly, to satisfy all these conditions, the quality of anchors as a positive sample needs to be evaluated reflecting the \textit{model's current learning status}, i.e. its parameter values.

With this motivation, we propose a probabilistic anchor assignment (PAA) strategy that adaptively separates a set of anchors into positive and negative samples for a GT box according to the learning status of the model associated with it. To do so we first define a score of a detected bounding box that reflects both the classification and localization qualities. We then identify the connection between this score and the training objectives and represent the score as the combination of two loss objectives. Based on this scoring scheme, we calculate the scores of individual anchors that reflect how the model finds useful cues to detect a target object in each anchor. With these anchor scores, we aim to find a probability distribution of two modalities that best represents the scores as positive or negative samples as in Figure \ref{fig1}. Under the found probability distribution, anchors with probabilities from the positive component are high are selected as positive samples. This transforms the anchor assignment problem to a maximum likelihood estimation for a probability distribution where the parameters of the distribution is determined by anchor scores. Based on the assumption that anchor scores calculated by the model are samples drawn from a probability distribution, it is expected that the model can infer the sample separation in a probabilistic way, leading to easier training of the model compared to other non-probabilistic assignments. Moreover, since positive samples are adaptively selected based on the anchor score distribution, it does not require a pre-defined number of positive samples nor an IoU threshold.

\begin{figure}[t]
	\begin{center}
		\includegraphics[width=1.0\linewidth]{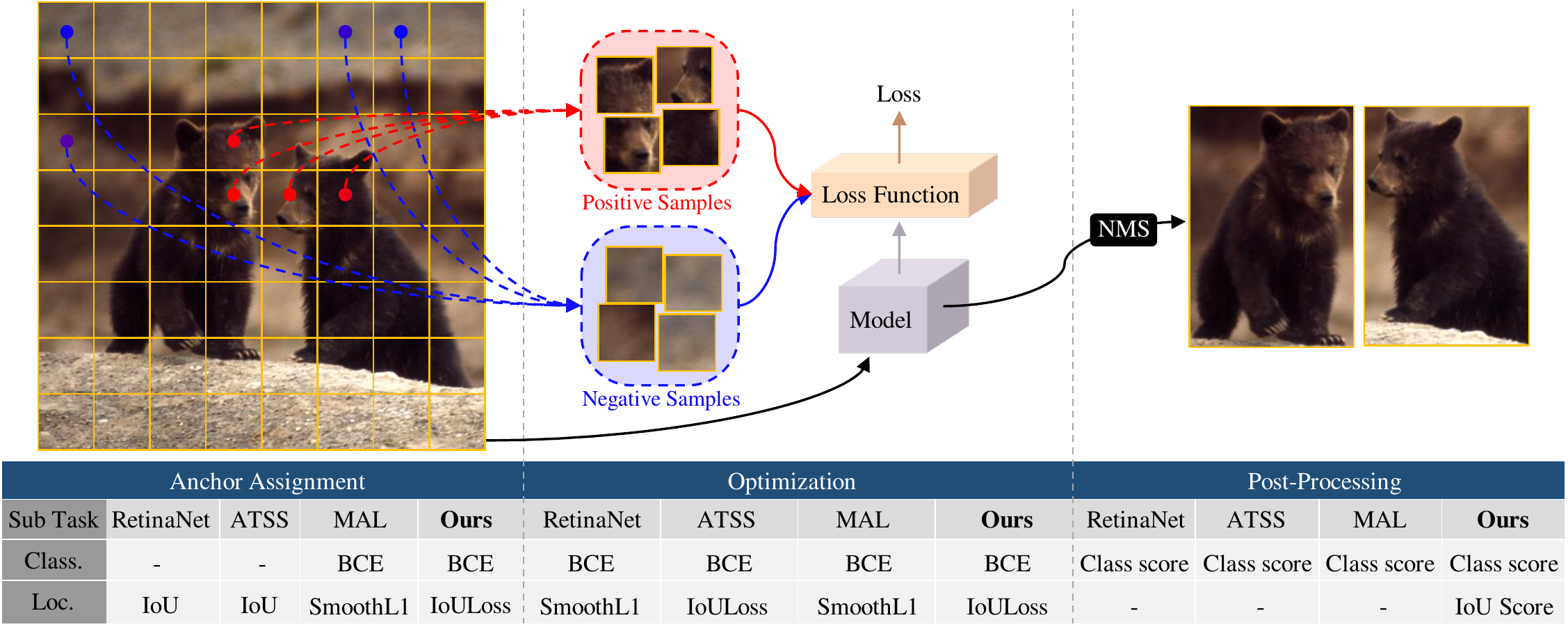}
	\end{center}
	\caption{Illustration of three key procedures of object detectors and comparison between RetinaNet\cite{focal}, ATSS\cite{atss}, MAL\cite{mal} and ours about in which form classification and localization tasks are concerned in each procedure. Unlike others, we take both tasks into account for all three procedures. For the localization task, we use IoU-based metrics to align the objectives of each procedure.}
	\label{fig0}
\end{figure}

On top of that, we identify that in most modern object detectors, there is inconsistency between the testing scheme (selecting boxes according to the classification score only during NMS) and the training scheme (minimizing both classification and localization losses). Ideally, the quality of detected boxes should be measured based not only on classification but also on localization. To improve this incomplete scoring scheme and at the same time to reduce the discrepancy of objectives between the training and testing procedures, we propose to predict the IoU of a detected box as a localization quality, and multiply the classification score by the IoU score as a metric to rank detected boxes. This scoring is intuitive, and allows the box scoring scheme in the testing procedure to share the same ground not only with the objectives used during training, but also with the proposed anchor assignment strategy that brings both classification and localization into account, as depicted in Figure \ref{fig0}.  Combined with the proposed PAA, this simple extension significantly contributes to detection performance. We also compare the IoU prediction with the centerness prediction\cite{fcos, atss} and show the superiority of the proposed method.

With an additional improvement in post-processing named score voting, each of our methods shows clear improvements as revealed in the ablation studies. In particular, on COCO test-dev set\cite{coco} all our models achieve new state-of-the-art performance with significant margins. Our model only requires to add a single convolutional layer, and uses a single anchor per spatial locations similar to \cite{atss}, resulting in a smaller number of parameters compared to RetinaNet\cite{focal}. The proposed anchor assignment can be parallelized using GPUs and does not require extra computes in testing time. All this evidence verifies the efficacy of our proposed methods. The contributions of this paper are summarized as below:

1. We model the anchor assignment as a probabilistic procedure by calculating anchor scores from a detector model and maximizing the likelihood of these scores for a probability distribution. This allows the model to infer the assignment in a probabilistic way and adaptively determines positive samples.

2. To align the objectives of anchor assignment, optimization and post-processing procedures, we propose to predict the IoU of detected boxes and use the unified score of classification and localization as a ranking metric for NMS. On top of that, we propose the score voting method as an additional post-processing using the unified score to further boost the performance.

3. We perform extensive ablation studies and verify the effectiveness of the proposed methods. Our experiments on MS COCO dataset with five backbones set up new AP records for all tested settings.

\section{Related Work}
\subsection{Recent Advances in Object Detection}
Since Region-CNN\cite{rcnn} and its improvements\cite{fastrcnn, fasterrcnn}, the concept of anchors and offset regression between anchors and ground truth (GT) boxes along with object category classification has been widely adopted. In many cases, multiple anchors of different scales and aspect ratios are assigned to each spatial location to cover various object sizes and shapes. Anchors that have IoU values greater than a threshold with one of GT boxes are considered as positive samples. Some systems use two-stage detectors\cite{rcnn, fastrcnn, fasterrcnn, fpn}, which apply the anchor mechanism in a region proposal network (RPN) for class-agnostic object proposals. A second-stage detection head is run on aligned features\cite{fasterrcnn, maskrcnn} of each proposal. Some systems use single-stage detectors\cite{yolo, ssd, yolo2, refinedet, focal, m2det}, which does not have RPN and directly predict object categories and regression offsets at each spatial location. More recently, anchor-free models that do not rely on anchors to define positive and negative samples and regression offsets have been introduced. These models predict various key points such as corners\cite{cornernet}, extreme points\cite{extremepoints}, center points\cite{objectsaspoints, fcos} or arbitrary feature points\cite{reppoints} induced from deformable convolution\cite{deformableconv}. \cite{fsaf} combines anchor-based detectors with anchor-free detection by adding additional anchor-free regression branches. It has been found in \cite{atss} that anchor-based and anchor-free models show similar performance when they use the same anchor assignment strategy.

\subsection{Anchor Assignment in Object Detection}
The task of selecting which anchors (or locations for anchor-free models) are to be designated as positive or negative samples has recently been identified as a crucial factor that greatly affects a model's performance\cite{metaanchor, freeanchor, atss}. In this regard, several methods have been proposed to overcome the limitation of the IoU-based hard anchor assignment. MetaAnchor\cite{metaanchor} predicts the parameters of the anchor functions (the last convolutional layers of detection heads) dynamically and takes anchor shapes as an argument, which provides the ability to change anchors in training and testing. Rather than enumerating pre-defined anchors across spatial locations, GuidedAnchoring\cite{guidedanchor} defines the locations of anchors near the center of GTs as positives and predicts their shapes. FreeAnchor\cite{freeanchor} proposes a detection-customized likelihood that considers both the recall and precision of samples into account and determines positive anchors based on the estimated likelihood. ATSS\cite{atss} suggests an adaptive anchor assignment that calculates the mean and standard deviation of IoU values from a set of close anchors for each GT. It assigns anchors whose IoU values are higher than the sum of the mean and the standard deviation as positives. Although these works show some improvements, they either require additional layers and complicated structures\cite{guidedanchor, metaanchor}, or force only one anchor to have a full classification score which is not desirable in cases where multiple anchors are of high quality and competitive\cite{freeanchor}, or rely on IoUs between pre-defined anchors and GTs and consider neither the actual content of the intersecting regions nor the model's learning status\cite{atss}.

Similar to our work, MultipleAnchorLearning (MAL)\cite{mal} and NoisyAnchor\cite{noisy} define anchor score functions based on classification and localization losses. However, they do not model the anchor selection procedure as a likelihood maximization for a probability distribution; rather, they choose a fixed number of best scoring anchors. Such a mechanism prevents these models from selecting a flexible number of positive samples according to the model's learning status and input. MAL uses a linear scheduling that reduces the number of positives as training proceeds and requires a heuristic feature perturbation to mitigate it. NoisyAnchor fixes the number of positive samples throughout training. Also, they either miss the relation between the anchor scoring scheme and the box selection objective in NMS\cite{mal} or only indirectly relate them using soft-labels\cite{noisy}.

\subsection{Predicting Localization Quality in Object Detection}
Predicting IoUs as a localization quality of detected bounding boxes is not new. YOLO and YOLOv2\cite{yolo, yolo2} predict ``objectness score'', which is the IoU of a detected box with its corresponding GT box, and multiply it with the classification score during inference. However, they do not investigate its effectiveness compared to the method that uses classification scores only, and their latest version \cite{yolov3} removes this prediction. IoU-Net\cite{iounet} also predicts the IoUs of predicted boxes and proposed ``IoU-guided NMS'' that uses predicted IoUs instead of classification scores as the ranking keyword, and adjusts the selected box's score as the maximum score of overlapping boxes. Although this approach can be effective, they do not correlate the classification score with the IoU as a unified score, nor do they relate the NMS procedure and the anchor assignment process.  In contrast to predicting IoUs, some works\cite{varvoting, gaussyolo} add an additional head to predict the variance of localization to regularize training\cite{varvoting} or penalize the classification score in testing\cite{gaussyolo}.

\section{Proposed Methods}
\subsection{Probabilistic Anchor Assignment Algorithm}
Our goal here is to devise an anchor assignment strategy that takes three key considerations into account: Firstly, it should measure the quality of a given anchor based on how likely the model associated with it finds evidence to identify the target object with that anchor. Secondly, the separation of anchors into positive and negative samples should be adaptive so that it does not require a hyperparameter such as an IoU threshold. Lastly, the assignment strategy should be formulated as a likelihood maximization for a probability distribution in order for the model to be able to reason about the assignment in a probabilistic way. In this respect, we design an anchor scoring scheme and propose an anchor assignment that brings the scoring scheme into account.

Specifically, let us define the score of an anchor that reflects the quality of its bounding box prediction for the closest ground truth (GT) $g$. One intuitive way is to calculate a classification score (compatibility with the GT class) and a localization score (compatibility with the GT box) and multiply them: 
\begin{equation}
  S(f_{\theta}(a, x), g) = S_{cls}(f_{\theta}(a, x), g) \times S_{loc}(f_{\theta}(a, x), g)^{\lambda}
\label{eq1}
\end{equation}
where $S_{cls}$, $S_{loc}$, and $\lambda$ are the score of classification and localization of anchor $a$ given $g$ and a scalar to control the relative weight of two scores, respectively. $x$ and $f_{\theta}$ are an input image and a model with parameters $\theta$. Note that this scoring function is dependent on the model parameters $\theta$. We can define and get $S_{cls}$ from the output of the classification head. How to define $S_{loc}$ is less obvious, since the output of the localization head is encoded offset values rather than a score. Here we use the Intersection-over-Union (IoU) of a predicted box with its GT box as $S_{loc}$, as its range matches that of the classification score and its values naturally correspond to the quality of localization:
\begin{equation}
  S_{loc}(f_{\theta}(a, x), g) = \mathrm{IoU}(f_{\theta}(a, x), g)
\end{equation}

Taking the negative logarithm of score function $S$, we get the following:
\begin{align}
\begin{split}
  -\log S(f_{\theta}(a, x), g) &= -\log S_{cls}(f_{\theta}(a, x), g) -\lambda\log S_{loc}(f_{\theta}(a, x), g) \\
  &= \mathcal{L}_{cls}(f_{\theta}(a, x), g) + \lambda\mathcal{L}_{IoU}(f_{\theta}(a, x), g)
  \label{eq:losssum}
\end{split}
\end{align}
where $\mathcal{L}_{cls}$ and $\mathcal{L}_{IoU}$ denote binary cross entropy loss\footnote{We assume a binary classification task. Extending it to a multi-class case is straightforward.} and IoU loss\cite{iouloss} respectively. One can also replace any of the losses with a more advanced objective such as Focal Loss\cite{focal} or GIoU Loss\cite{giou}. It is then legitimate that the negative sum of the two losses can act as a scoring function of an anchor given a GT box.

To allow a model to be able to reason about whether it should predict an anchor as a positive sample in a probabilistic way, we model anchor scores for a certain GT as samples drawn from a probability distribution and maximize the likelihood of the anchor scores w.r.t the parameters of the distribution. The anchors are then separated into positive and negative samples according to the probability of each being a positive or a negative. Since our goal is to distinguish a set of anchors into two groups (positives and negatives), any probability distribution that can model the multi-modality of samples can be used. Here we choose Gaussian Mixture Model (GMM) of two modalities to model the anchor score distribution. 
\begin{align}
P(a|x,g,\theta) = w_{1}\mathcal{N}_{1}(a;m_1,p_1) + w_{2}\mathcal{N}_{2}(a;m_2,p_2)
\end{align}
where $w_1, m_1, p_1$ and $w_2, m_2, p_2$ represent the weight, mean and precision of two Gaussians, respectively. Given a set of anchor scores, the likelihood of this GMM can be optimized using Expectation-Maximization (EM) algorithm.

\begin{figure}[t]
	\begin{center}
		\includegraphics[width=1.0\linewidth]{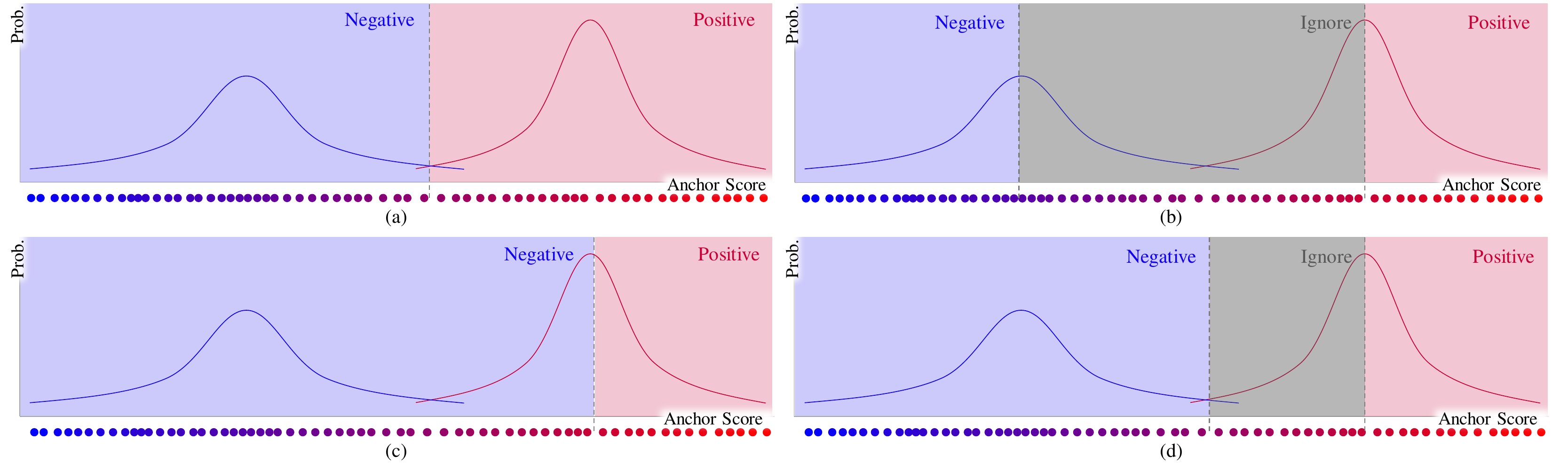}
	\end{center}
	\caption{Different boundary schemes to separate anchors using their probabilities.}
	\label{fig2}
\end{figure}

With the parameters of GMM estimated by EM, the probability of each anchor being a positive or a negative sample can be determined. With these probability values, various techniques can be used to separate the anchors into two groups. Figure \ref{fig2} illustrates different examples of separation boundaries based on anchor probabilities. The proposed algorithm using one of these boundary schemes is described in Procedure \ref{alg1}.  To calculate anchor scores, anchors are first allocated to the GT of the highest IoU (Line 3). To make EM efficient, we collect top $K$ anchors from each pyramid level (Line 5-11) and perform EM (Line 12). Non-top $K$ anchors are assigned as negative samples (Line 16). 

Note that the number of positive samples is adaptively determined depending on the estimated probability distribution conditioned on the model's parameters. This is in contrast to previous approaches that ignore the model\cite{atss} or heuristically determine the number of samples as a hyperparameter\cite{mal, noisy} without modeling the anchor assignment as a likelihood maximization for a probability distribution. FreeAnchor\cite{freeanchor} defines a detection-customized likelihood and models the product of the recall and the precision as the training objective. But their approach is significantly different than ours in that we do not separately design likelihoods for recall and precision, nor do we restrict the number of anchors that have a full classification score to one. In contrast, our likelihood is based on a simple one-dimensional GMM of two modalities conditioned on the model's parameters, allowing the anchor assignment strategy to be easily identified by the model. This results in easier learning compared to other anchor assignment methods that require complicated sub-routines (e.g. the mean-max function to stabilize training\cite{freeanchor} or the anchor depression procedure to avoid local minima\cite{mal}) and thus leads to better performance as shown in the experiments.

\begin{algorithm} [tb]
	\caption{Probabilistic anchor assignment algorithm.}
	\label{alg1}
	\begin{algorithmic}[1]
		\REQUIRE $\mathcal{G}$, $\mathcal{A}$, $\mathcal{A}_i$, $\mathcal{L}$, $\mathcal{K}$ \newline $\mathcal{G}$ is a set of ground-truth boxes \newline $\mathcal{A}$ is a set of all anchor boxes \newline $\mathcal{A}_i$ is a set of anchor boxes from $i_{th}$ pyramid level \newline $\mathcal{L}$ is the number of pyramid levels \newline $\mathcal{K}$ is the number of candidate anchors for each pyramid
		\ENSURE $\mathcal{P}$, $\mathcal{N}$, $\mathcal{I}$ \newline $\mathcal{P}$ is a set of positive samples \newline $\mathcal{N}$ is a set of negative samples \newline
		$\mathcal{I}$ is a set of ignoring samples
		\STATE $\mathcal{P} \leftarrow \varnothing, \mathcal{N} \leftarrow \varnothing$
		\FOR{$g \in \mathcal{G}$}
		    \STATE $\mathcal{A}_g \leftarrow$ GetAnchors($\mathcal{A}$, $g$, $\mathcal{G}$) \COMMENT {Get all anchors that has $g$ as best GT w.r.t. IoU.}
		    \STATE $\mathcal{C}_g \leftarrow \varnothing$
    		\FOR{$i = 1$ \TO $\mathcal{L}$}
    		\STATE $\mathcal{A}_i^g \leftarrow \mathcal{A}_i \cap \mathcal{A}_g$
    		\STATE $\mathcal{S}_i \leftarrow$ ComputeAnchorScores($\mathcal{A}_i^g$, $g$) \COMMENT{Negative of Eq.\ref{eq:losssum}}
    		\STATE $t_i \leftarrow$ FindKthLargest($s_i, \mathcal{K}$)
    		\STATE $\mathcal{C}_g^i \leftarrow \{ a_{j} \in \mathcal{A}_i^g \mid t_i \leq s_{j} \in \mathcal{S}_i\}$
    		\STATE $\mathcal{C}_g \leftarrow \mathcal{C}_g \cup \mathcal{C}_g^i$
    		\ENDFOR
    	\STATE $\mathcal{B}, \mathcal{F} \leftarrow$ FitGMM($\mathcal{C}_g$, 2) \COMMENT{$\mathcal{B}$, $\mathcal{F}$: Probabilties of two Gaussians for $\mathcal{C}_g$}
    	\STATE $\mathcal{N}_g, \mathcal{P}_g \leftarrow \text{SeparateAnchors}(\mathcal{C}_g, B, F)$ \COMMENT{Separate anchors using one of Fig.\ref{fig2}.}
    	\STATE $\mathcal{P} \leftarrow \mathcal{P} \cup \mathcal{P}_g, \mathcal{N}\leftarrow \mathcal{N}\cup \mathcal{N}_g, \mathcal{I} \leftarrow \mathcal{I} \cup (\mathcal{C}_g - \mathcal{P}_g-\mathcal{N}_g)$
		\ENDFOR
		\STATE $\mathcal{N} \leftarrow \mathcal{N} \cup (\mathcal{A} - \mathcal{P}-\mathcal{N}-\mathcal{I})$
		\RETURN $\mathcal{P}$, $\mathcal{N}$, $\mathcal{I}$
	\end{algorithmic}
\end{algorithm}

To summarize our method and plug it into the training process of an object detector, we formulate the final training objective for an input image $x$ (we omit $x$ for brevity):
\begin{equation}
  \operatorname*{argmax}_{\theta} \prod_g\prod_{a \in \mathcal{A}_g} P_{pos}(a, \theta, g)S_{pos}(a, \theta, g) + P_{neg}(a, \theta, g)S_{neg}(a, \theta)
\label{eq:objective}
\end{equation}
\begin{align}
\begin{split}
  S_{pos}(a, \theta, g) &= S(f_{\theta}(a), g) \\
  &= \operatorname{exp}(-\mathcal{L}_{cls}(f_{\theta}(a), g) - \lambda\mathcal{L}_{IoU}(f_{\theta}(a)), g)
\end{split}
\end{align}
\begin{equation}
  S_{neg}(a, \theta) = \operatorname{exp}(-\mathcal{L}_{cls}(f_{\theta}(a), \varnothing))
\end{equation}
where $P_{pos}(a, \theta, g)$ and $P_{neg}(a, \theta, g)$ indicate the probability of an anchor being a positive or a negative and can be obtained by the proposed PAA. $\varnothing$ means the background class. Our PAA algorithm can be viewed as a procedure to compute $P_{pos}$ and $P_{neg}$ and approximate them as binary values (i.e. separate anchors into two groups) to ease optimization. In each training iteration, after estimating $P_{pos}$ and $P_{neg}$, the gradients of the loss objectives w.r.t. $\theta$ can be calculated and stochastic gradient descent can be performed.

\subsection{IoU Prediction as Localization Quality}
The anchor scoring function in the proposed anchor assignment is derived from the training objective (i.e. the combined loss of two tasks), so the anchor assignment procedure is well aligned with the loss optimization. However, this is not the case for the testing procedure where the non-maximum suppression (NMS) is performed solely on the classification score. To remedy this, the localization quality can be incorporated into NMS procedure so that the same scoring function (Equation \ref{eq1}) can be used. However, GT information is only available during training, and so IoU between a detected box and its corresponding GT box cannot be computed at test time.

Here we propose a simple solution to this: we extend our model to predict the IoU of a predicted box with its corresponding GT box. This extension is straightforward as it requires a single convolutional layer as an additional prediction head that outputs a scalar value per anchor. We use Sigmoid activation on the output to obtain valid IoU values. The training objective then becomes (we omit input x for brevity):
\begin{align}
    \mathcal{L}(f_{\theta}(a), g) = \mathcal{L}_{cls}(f_{\theta}(a), g) + \lambda_1\mathcal{L}_{IoU}(f_{\theta}(a), g) + \lambda_2\mathcal{L}_{IoUP}(f_{\theta}(a), g)
\label{eq:loss}
\end{align}
where ${L}_{IoUP}$ is IoU prediction loss defined as binary cross entropy between predicted IoUs and true IoUs. With the predicted IoU, we compute the unified score of the detected box using Equation \ref{eq1} and use it as a ranking metric for NMS procedure. As shown in the experiments, bringing IoU prediction into NMS significantly improves performance, especially when coupled with the proposed probabilistic anchor assignment. The overall network architecture is exactly the same as the one in FCOS\cite{fcos} and ATSS\cite{atss}, which is RetinaNet with modified feature towers and an auxiliary prediction head. Note that this structure uses only a single anchor per spatial location and so has a smaller number of parameters and FLOPs compared to RetinaNet-based models using nine anchors.

\subsection{Score Voting}
As an additional improvement method here we propose a simple yet effective post-processing scheme. The proposed score voting method works on each box $b$ of remaining boxes after NMS procedure as follows:
\begin{equation}
    \begin{split}
        p_i = e^{-(1-\text{IoU}(b, b_i))^2/\sigma_t}
    \end{split}
\end{equation}
\begin{equation}
    \begin{split}
        \hat{b}=\dfrac{\sum_ip_is_ib_i}{\sum_ip_is_i} \text{ subject to IoU}(b, b_i) > 0
    \end{split}
\end{equation}
where $\hat{b}$, $s_i$ and $\sigma_t$ is the updated box, the score computed by Equation \ref{eq1} and a hyperparameter to adjust the weights of adjacent boxes $b_i$ respectively. It is noted that this voting algorithm is inspired by ``variance voting'' described in \cite{varvoting} and $p_i$ is defined in the same way. However, we do not use the variance prediction to calculate the weight of each neighboring box. Instead we use the unified score of classification and localization $s_i$ as a weight along with $p_i$.

We found that using $p_i$ alone as a box weight leads to a performance improvement, and multiplying it by $s_i$ further boost the performance. In contrast to the variance voting, detectors without the variance prediction are capable of using the score voting by just weighting boxes with $p_i$. Detectors with IoU prediction head, like ours, can multiply it by $s_i$ for better accuracy. Unlike the classification score only, $s_i$ can act as a reliable weight since it does not assign large weights to boxes that have a high classification score and a poor localization quality.

\section{Experiments}
\label{experiments}
In this section we conduct extensive experiments to verify the effectiveness of the proposed methods on MS COCO benchmark\cite{coco}. We follow the common practice of using `trainval35k' as training data (about 118k images) for all experiments. For ablation studies we measure accuracy on `minival' of 5k images and comparisons with previous methods are done on `test-dev' of about 20k images. All accuracy numbers are computed using the official COCO evaluation code.

\subsection{Training Details}
We use a COCO training setting which is the same as \cite{atss} in the batch size, frozen Batch Normalization, learning rate, etc. The exact setting can be found in the supplementary material. For ablation studies we use Res50 backbone and run 135k iterations of training. For comparisons with previous methods we run 180k iterations with various backbones. Similar to recent works\cite{fcos, atss}, we use GroupNorm\cite{groupnorm} in detection feature towers, Focal Loss\cite{focal} as the classification loss, GIoU Loss\cite{giou} as the localization loss, and add trainable scalars to the regression head. $\lambda_1$ is set to 1 to compute anchor scores and 1.3 when calculating Equation \ref{eq:losssum}. $\lambda_2$ is set to 0.5 to balance the scales of each loss term. $\sigma_t$ is set to 0.025 if the score voting is used. Note that we do \textit{not} use ``centerness'' prediction or ``center sampling''\cite{fcos, atss} in our models. We set $\mathcal{K}$ to 9 although our method is not sensitive to its value similar to \cite{atss}. For GMM optimization, we set the minimum and maximum score of the candidate anchors as the mean of two Gaussians and set the precision values to one as an initialization of EM.
\subsection{Ablation Studies}
\begin{table}[t]
\begin{center}
\caption {Ablation studies on COCO minival set with Res50 backbone. \textbf{Left}: Comparison of anchor separation boundaries in Figure \ref{fig2}, fixed numbers of positives (FNP) and fixed positive score ranges (FSR). \textbf{Right}: Effects of individual methods.}
\begin{minipage}{.4\linewidth}
\centering
\begin{tabular}{c|ccc}
\hline\noalign{\smallskip}
Anchor Sep. & AP & AP50 & AP75 \\
\noalign{\smallskip}
\hline
 Fig.\ref{fig2}. (a) & 40.5 & 58.8 & 43.4\\
 Fig.\ref{fig2}. (b) & 40.5 & 58.9 & 43.8\\
 Fig.\ref{fig2}. (c) & 40.9 & 59.4 & 43.9\\
 Fig.\ref{fig2}. (d) & 40.7 & 59.1 & 44.0\\
 \hline
 FNP (5) & 39.5 & 58.0 & 42.7\\
 FNP (10) & 40.1 & 58.5 & 43.3\\
 FNP (20) & 40.0 & 58.5 & 43.1\\
 FSR ($>$ 0.1) & 23.8 & 38.5 & 25.2\\
 FSR ($>$ 0.2) & 19.3 & 33.2 & 19.8\\
 FSR ($>$ 0.3) &\multicolumn{3}{c}{training failed} \\
\hline
\end{tabular}
\label{table:sep}
\end{minipage}%
\begin{minipage}{.6\linewidth}
\centering
\begin{tabular}{ccc|ccc}
\hline\noalign{\smallskip}
Method & Aux. task & Voting & AP & AP50 & AP75 \\
\noalign{\smallskip}
\hline
 IoU & & & 34.6 & 53.0 & 36.7\\
 IoU& IoU pred. & & 36.0 & 54.0 & 38.9\\
PAA & & & 39.9 & 59.1 & 42.8\\
PAA & Center pred. & & 39.8 & 58.3 & 43.2\\
PAA & IoU pred. & & 40.9 & 59.4 & 43.9\\
PAA & IoU pred. &\checkmark & \textbf{41.1} & \textbf{59.4} & \textbf{44.3} \\
ATSS & Center pred. & & 39.4 & 57.4 & 42.4\\
ATSS & IoU pred. & & 39.8 & 57.9 & 43.2\\
\hline
\label{table:ablation}
\end{tabular}
\end{minipage}%
\end{center}
\end{table}
\begin{table}[t]
\begin{center}
\caption {\textbf{Left}: Performance comparison on COCO minival dataset with Res50 backbone. All models were trained with 135K iterations. \textbf{Right}: Average errors of IoU prediction on COCO minival set for various backbones.}
\begin{minipage}{.6\linewidth}
\centering
\begin{tabular}{c|cccccc}
\hline\noalign{\smallskip}
Method & AP & AP50 & AP75 & APs & APm & APl \\
\noalign{\smallskip}
\hline
RetinaNet & 36.7 & 55.8 & 39.0 & 19.7 & 40.1 & 49.1\\
MAL\cite{mal} &  38.4 & 56.8 & 41.1 & - & - & -\\
ATSS\cite{atss} & 39.4 & 57.4 & 42.4 & 23.0 & 42.9 & 51.9\\
Ours & \textbf{41.1} & \textbf{59.4} & \textbf{44.3} & \textbf{23.5} & \textbf{45.4} & \textbf{54.3}\\
\hline
\end{tabular}
\label{table:val}
\end{minipage}%
\begin{minipage}{.4\linewidth}
\centering
\begin{tabular}{c|c}
\hline\noalign{\smallskip}
Backbone & IoU Pred. Err.\\
\noalign{\smallskip}
\hline
Res50 & 0.093\\
Res101 & 0.092\\
ResNext101 & 0.09\\
Res101-DCN & 0.086\\
\hline
\end{tabular}
\label{table:iou}
\end{minipage}%
\end{center}
\end{table}

\subsubsection{Comparison between different anchor separation points}
Here we compare the anchor separation boundaries depicted in Figure \ref{fig2}. The left table in Table \ref{table:ablation} shows the results. All the separation schemes work well, and we find that (c) gives the most stable performance. We also compare our method with two simpler methods, namely fixed numbers of positives (FNP) and fixed positive score ranges (FSR). FNP defines a pre-defined number of top-scoring samples as positives while FSR treats all anchors whose scores exceed a certain threshold as positives. As the results in the right of Table \ref{table:ablation} show, both methods show worse performance than PAA. FSR ($>$ 0.3) fails because the model cannot find anchors whose scores are within the range at early iterations. This shows an advantage of PAA that adaptively determines the separation boundaries without hyperparameters that require careful hand-tuning and so are hard to be adaptive per data.

\subsubsection{Effects of individual modules}
In this section we verify the effectiveness of individual modules of the proposed methods. Accuracy numbers for various combinations are in Table \ref{table:ablation}. Changing anchor assignment from the IoU-based hard assignment to the proposed PAA shows improvements of 5.3\% in AP score. Adding IoU prediction head and applying the unified score function in NMS procedure further boosts the performance to 40.8\%. To further verify the impact of IoU prediction, we compare it with centerness prediction used in \cite{fcos, atss}. As can be seen in the results, centerness does not bring improvements to PAA. This is expected as weighting scores of detected boxes according to its centerness can hinder the detection of acentric or slanted objects. This shows that centerness-based scoring does not generalize well and the proposed IoU-based scoring can overcome this limitation. We also verify that IoU prediction is more effective than centerness prediction for ATSS\cite{atss} (39.8\% vs. 39.4\%).  Finally, applying the score voting improves the performance to 41.0\%, surpassing previous methods with Res50 backbone in Table \ref{table:val}.Left with significant margins.

\begin{figure}[t]
    \begin{subfigure}{.65\textwidth}
		\includegraphics[width=.95\linewidth]{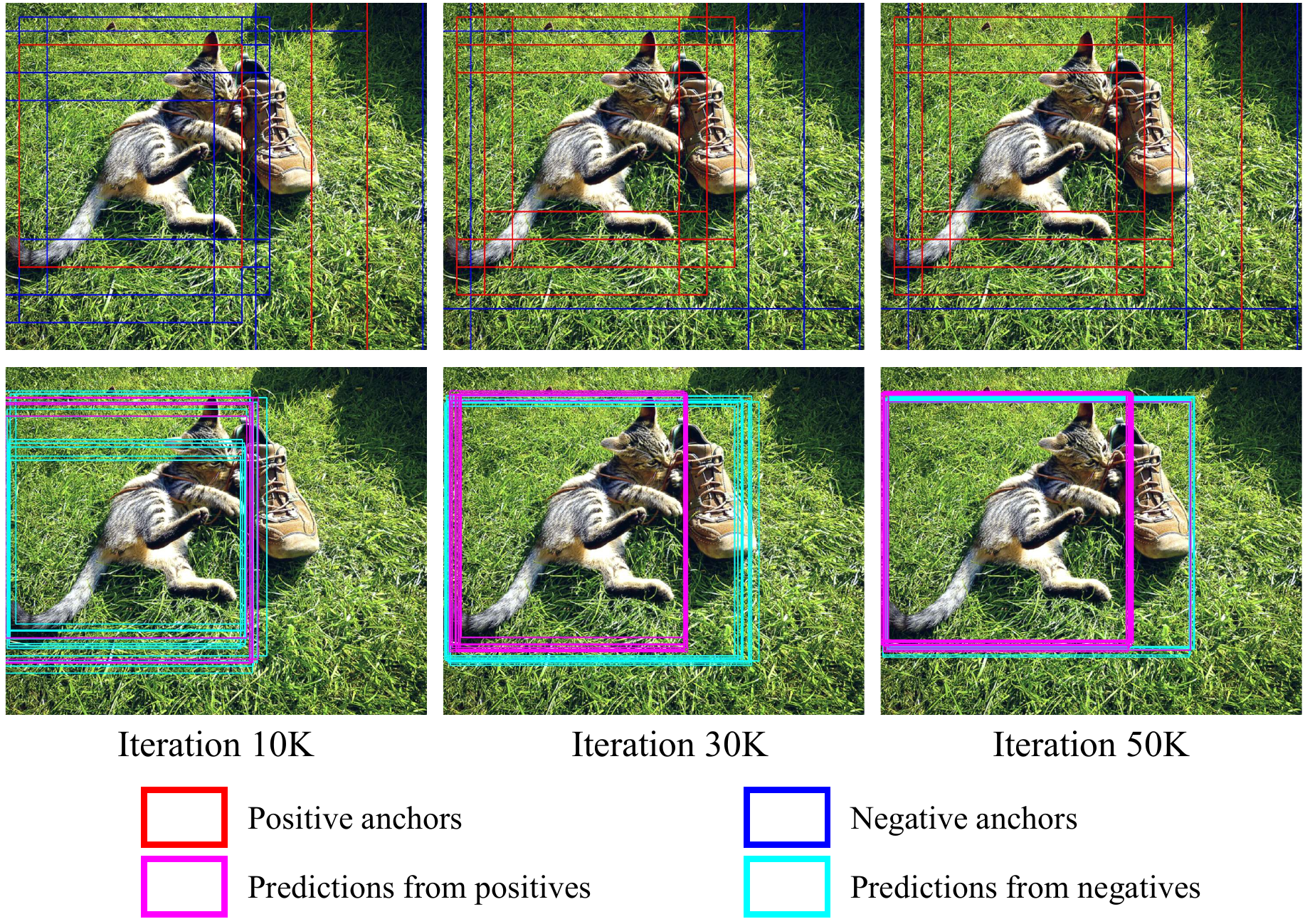}
	    \caption{}
	    \label{fig_box_vis}
    \end{subfigure}
    \begin{subfigure}{.34\textwidth}
		\includegraphics[width=0.98\linewidth]{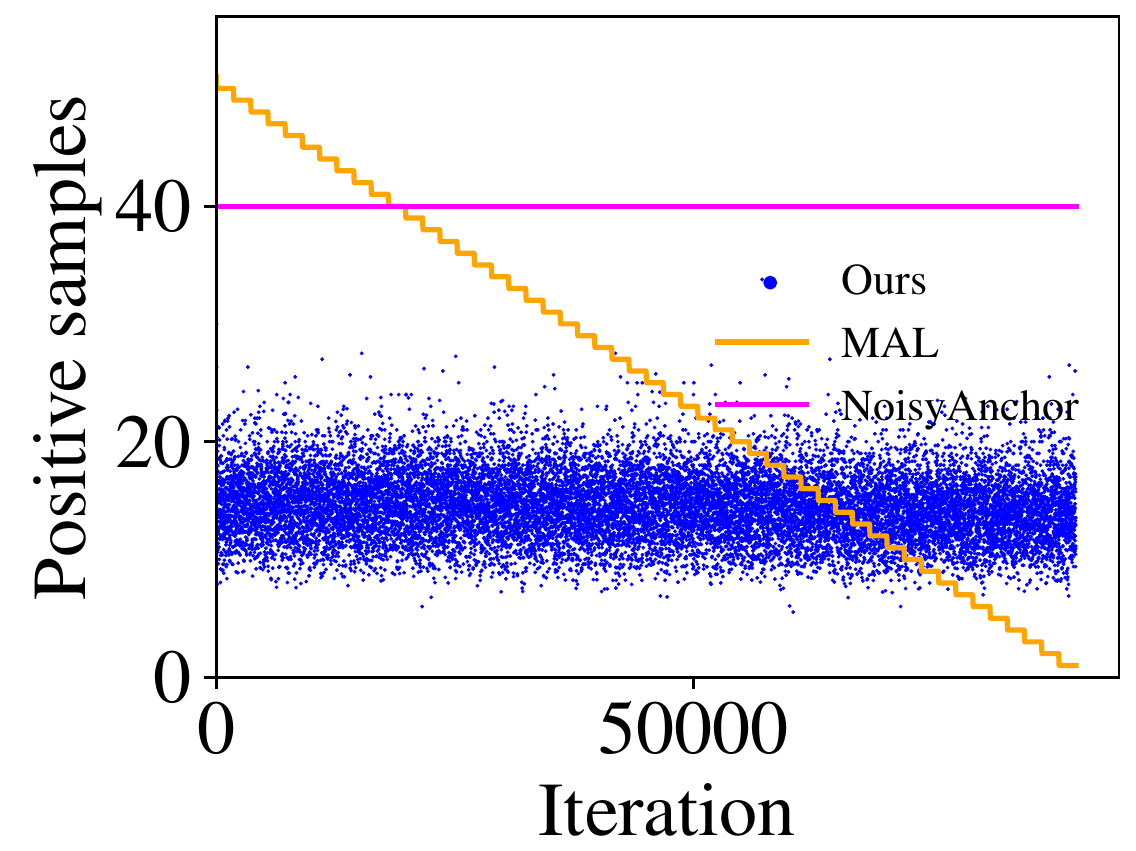}
		\caption{}
	    \label{fig_numpos}
    \end{subfigure}
\caption{(a) Evolution of anchor assignment and predicted boxes during training. (b) Plot of the number of positive samples per single GT box throughout training iterations. For our method, the numbers are averaged over a GPU for better visualization (individual values vary between 1 and 40).}
\end{figure}

\subsubsection{Accuracy of IoU prediction}
We calculate the average error of IoU prediction for various backbones in Table \ref{table:iou}.Right. All backbones show less than 0.1 errors, showing that IoU prediction is plausible with an additional convolutional head.

\subsubsection{Visualization of anchor assignment}
We visualize positive and negative samples separated by PAA in Figure \ref{fig_box_vis}. As training proceeds, the distinction between positive and negative samples becomes clearer. Note that the positive anchors do not necessarily have larger IoU values with the target bounding box than the negative ones. Also, many negative anchors in the iteration 30k and 50k have high IoU values. Methods with a fixed number of positive samples\cite{mal, noisy} can assign these anchors as positives, and the model might predict these anchors with high scores during inference. Finally, many positive anchors have more accurate localization as training proceeds. In contrast to ours, methods like FreeAnchor\cite{freeanchor} penalize all these anchors except the single best one, which can confuse training. 

\subsubsection{Statistics of positive samples}
To compare our method and recent works that also select positive samples by scoring anchors, we plot the number of positive samples according to training iterations in Figure \ref{fig_numpos}. Unlike methods that either fix the number of samples \cite{noisy} or use a linear decay\cite{mal}, ours choose a different number of samples per iteration, showing the adaptability of the method.

\subsection{Comparison with State-of-the-art Methods}

\begin{table}[t]
\begin{center}
\caption {Results on COCO test-dev set. * indicates multi-scale testing. Bold text means the best performance among models with the same or a similar backbone.}
\begin{tabular}{c|c|ccc|ccc}
\hline
Method & Backbone & AP & AP50 & AP75 & APs & APm & APl \\
\hline
RetinaNet\cite{focal} & ResNet101 & 39.1 & 59.1 & 42.3 & 21.8 & 42.7 & 50.2\\
FCOS\cite{fcos} & ResNet101 & 41.5 & 60.7 & 45.0 & 24.4 & 44.8 & 51.6\\
NoisyAnchor\cite{noisy} &  ResNet101 & 41.8 & 61.1 & 44.9 & 23.4 & 44.9 & 52.9\\
FreeAnchor\cite{freeanchor} & ResNet101 & 43.1 & 62.2 & 46.4 & 24.5 & 46.1 & 54.8\\
MAL\cite{mal} &  ResNet101 & 43.6 & 61.8 & 47.1 & 25.0 & 46.9 & 55.8\\
ATSS\cite{atss} & ResNet101 & 43.6 & 62.1 & 47.4 & 26.1 & 47.0 & 53.6 \\
Ours & ResNet101 & \textbf{44.8} & \textbf{63.3} & \textbf{48.7} & \textbf{26.5} & \textbf{48.8} & \textbf{56.3} \\
\hline
FCOS\cite{fcos} & ResNeXt-64x4d-101 & 43.2 & 62.8 & 46.6 & 26.5 & 46.2 & 53.3 \\
NoisyAnchor\cite{noisy} & ResNeXt101 & 44.1 & 63.8 & 47.5 & 26.0 & 47.4 & 55.0 \\
FreeAnchor\cite{freeanchor} & ResNeXt-64x4d-101 & 44.9 & 64.3 & 48.5 & 26.8 & 48.3 & 55.9 \\
ATSS\cite{atss} & ResNeXt-64x4d-101 & 45.6 & 64.6 & 49.7 & 28.5 & 48.9 & 55.6 \\
MAL\cite{mal} & ResNeXt101 & 45.9 & 65.4 & 49.7 & 27.8 & 49.1 & 57.8 \\
Ours & ResNeXt-64x4d-101 & \textbf{46.6} & \textbf{65.6} & \textbf{50.8} & \textbf{28.8} & \textbf{50.4} & \textbf{57.9} \\
\hline
MAL\cite{mal}* & ResNeXt101 & 47.0 & 66.1 & 51.2 & 30.2 & 50.1 & 58.9 \\
Ours* & ResNeXt-64x4d-101 & \textbf{49.4} & \textbf{67.7} & \textbf{54.9} & \textbf{32.7} & \textbf{51.9} & \textbf{60.9} \\
\hline
RepPoints\cite{reppoints} & ResNet101-DCN & 45.0 & \textbf{66.1} & 49.0 & 26.6 & 48.6 & 57.5 \\
ATSS\cite{atss} & ResNet101-DCN & 46.3 & 64.7 & 50.4 & 27.7 & 49.8 & 58.4 \\
Ours & ResNet101-DCN & \textbf{47.4} & 65.7 & \textbf{51.6} & \textbf{27.9} & \textbf{51.3} & \textbf{60.6} \\
\hline
ATSS\cite{atss} & ResNeXt-64x4d-101-DCN & 47.7 & 66.5 & 51.9 & 29.7 & 50.8 & 59.4 \\
Ours & ResNeXt-64x4d-101-DCN & \textbf{49.0} & \textbf{67.8} & \textbf{53.3} & \textbf{30.2} & \textbf{52.8} & \textbf{62.2} \\
\hline
ATSS\cite{atss}* & ResNeXt-64x4d-101-DCN & 50.7 & 68.9 & 56.3 & 33.2 & 52.9 & 62.2 \\
Ours* & ResNeXt-64x4d-101-DCN & \textbf{51.4} & \textbf{69.7} & \textbf{57.0} & \textbf{34.0} & \textbf{53.8} & \textbf{64.0} \\
\hline
Ours & ResNeXt-32x8d-152-DCN & \textbf{50.8} & \textbf{69.7} & \textbf{55.1} & \textbf{31.4} & \textbf{54.7} & \textbf{65.2} \\
\hline
Ours* & ResNeXt-32x8d-152-DCN & \textbf{53.5} & \textbf{71.6} & \textbf{59.1} & \textbf{36.0} & \textbf{56.3} & \textbf{66.9} \\
\hline
\end{tabular}
\label{table:test}
\end{center}
\end{table}

To verify our methods with previous state-of-the-art ones, we conduct experiments with five backbones as in Table \ref{table:test}. We first compare our models trained with Res10 and previous models trained with the same backbone. Our Res101 model achieves 44.8\% accuracy, surpassing previous best models\cite{atss, mal} of 43.6 \%. With ResNext101 our model improves to 46.6\% (single-scale testing) and 49.4\% (multi-scale testing) which also beats the previous best model of 45.9\% and 47.0\%\cite{mal}. Then we extend our models by applying the deformable convolution to the backbones and the last layer of feature towers same as\cite{atss}. These models also outperform the counterparts of ATSS, showing 1.1\% and 1.3\% improvements. Finally, with the deformable ResNext152 backbone, our models set new records for both the single scale testing (50.8\%)  and the multi-scale testing (53.5\%).

\section{Conclusions}
In this paper we proposed a probabilistic anchor assignment (PAA) algorithm in which the anchor assignment is performed as a likelihood optimization for a probability distribution given anchor scores computed by the model associated with it. The core of PAA is in determining positive and negative samples in favor of the model so that it can infer the separation in a probabilistically reasonable way, leading to easier training compared to the heuristic IoU hard assignment or non-probabilistic assignment strategies. In addition to PAA, we identified the discrepancy of objectives in key procedures of object detection and proposed IoU prediction as a measure of localization quality to apply a unified score of classification and localization to NMS procedure. We also provided the score voting method which is a simple yet effective post-processing scheme that is applicable to most dense object detectors. Experiments showed that the proposed methods significantly boosted the detection performance, and surpassed all previous methods on COCO test-dev set.

\clearpage
%
%
\bibliographystyle{splncs04}
\bibliography{egbib}
\clearpage

\newpage
\input{supplementary}
\end{document}

%% file: supplementary.tex
\section{Appendix}

\subsection{Training Details}
We train our models with 8 GPUs each of which holds two images during training. The parameters of Batch Normalization layers\cite{batchnorm} are frozen as is a common practice. All backbones are pre-trained with ImageNet dataset\cite{imagenet}. We set the initial learning rate to 0.01 and decay it by a factor of 10 at 90k and 120k iterations for the 135k setting and at 120k and 160k for the 180k setting. For the 180k setting the multi-scale training strategy (resizing the shorter side of input images to a scale randomly chosen from 640 to 800) is adopted as is also a common practice. The momentum and weight decay are set to 0.9 and 1e-4 respectively. Following \cite{imagenet1hour} we use the learning rate warmup for the first 500 iterations. It is noted that multiplying individual localization losses by the scores of an auxiliary task (in our case, this is predicted IoUs with corresponding GT boxes, and centerness scores when using the centerness prediction as in \cite{fcos, atss}), which is also applied in previous works\cite{fcos, atss}, helps train faster and leads to a better performance.

\subsection{Network architecture}
\begin{figure}[t]
	\begin{center}
		\includegraphics[width=1.0\linewidth]{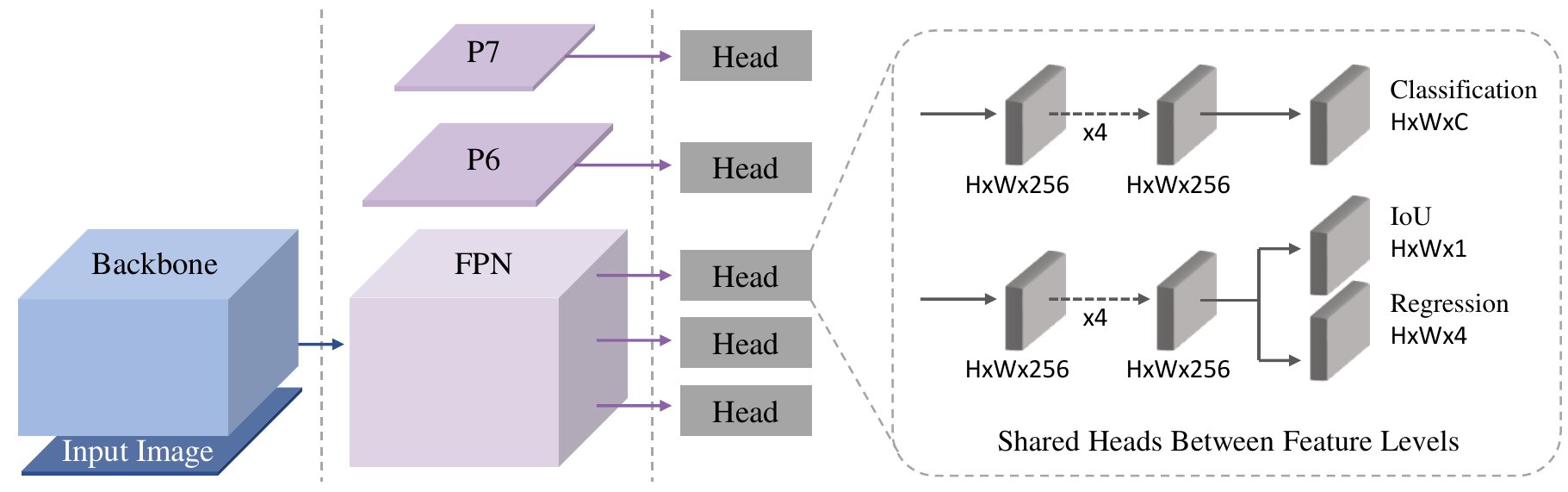}
	\end{center}
	\caption{The proposed model architecture. It has the same structure as FCOS\cite{fcos} and ATSS\cite{atss} but predicts the IoU of detected boxes instead of the centerness.}
	\label{fig3}
\end{figure}

Here we provide Figure \ref{fig3} for a visualization of our network architecture. It is a modified RetinaNet architecture with a single anchor per spatial location which is exactly the same as models used in FCOS\cite{fcos} and ATSS\cite{atss}. The only difference is that the additional head in our model predicts IoUs of predicted boxes whereas FCOS and ATTS models predict centerness scores.

\subsection{More Ablation Studies}
We conduct additional ablation studies regarding the effects of topk $\mathcal{K}$ and the default anchor scale. All the experiments in the main paper are conducted with $\mathcal{K}=9$ and the default anchor scale of 8. The anchor size for each pyramid level is determined by the product of its stride and the default anchor scale\footnote{So with the default anchor scale 8 and a feature pyramid of strides from 8 to 128, the anchor sizes are from 64 to 1024.}. Table \ref{table:extra_ablation_topk} shows the results on different default anchor scales. It shows that the proposed probabilistic anchor assignment is robust to both $\mathcal{K}$ and anchor sizes. 

\begin{table}
\begin{center}
\caption {Ablation studies on COCO minival set with Res50 backbone. \textbf{Left}: Comparison of different topk $\mathcal{K}$ values. \textbf{Right}: Comparison of different default anchor scales.}
\begin{minipage}{.4\linewidth}
\centering
\begin{tabular}{c|ccc}
\hline\noalign{\smallskip}
topk $\mathcal{K}$ & AP & AP50 & AP75 \\
\noalign{\smallskip}
\hline
 5 & 40.5 & 58.9 & 43.5\\
 9 & 40.9 & 59.4 & 43.9\\
 18 & 40.4 & 58.7 & 43.5\\
 25 & 40.7 & 58.9 & 43.9\\
\hline
\end{tabular}
\label{table:extra_ablation_topk}
\end{minipage}%
\begin{minipage}{.6\linewidth}
\centering
\begin{tabular}{c|ccc}
\hline\noalign{\smallskip}
default anchor scale & AP & AP50 & AP75 \\
\noalign{\smallskip}
\hline
 4 & 40.8 & 59.1 & 44.0\\
 6 & 40.7 & 59.5 & 43.8\\
 8 & 40.9 & 59.4 & 43.9\\
 10 & 40.8 & 59.3 & 43.9\\
\hline
\end{tabular}
\label{table:extra_ablation_default_anchor_scale}
\end{minipage}%
\end{center}
\end{table}

\subsection{More Visualization of Anchor Assignment}
\begin{figure}[t]
    \centering
    \begin{subfigure}{.85\textwidth}
		\includegraphics[width=.95\linewidth]{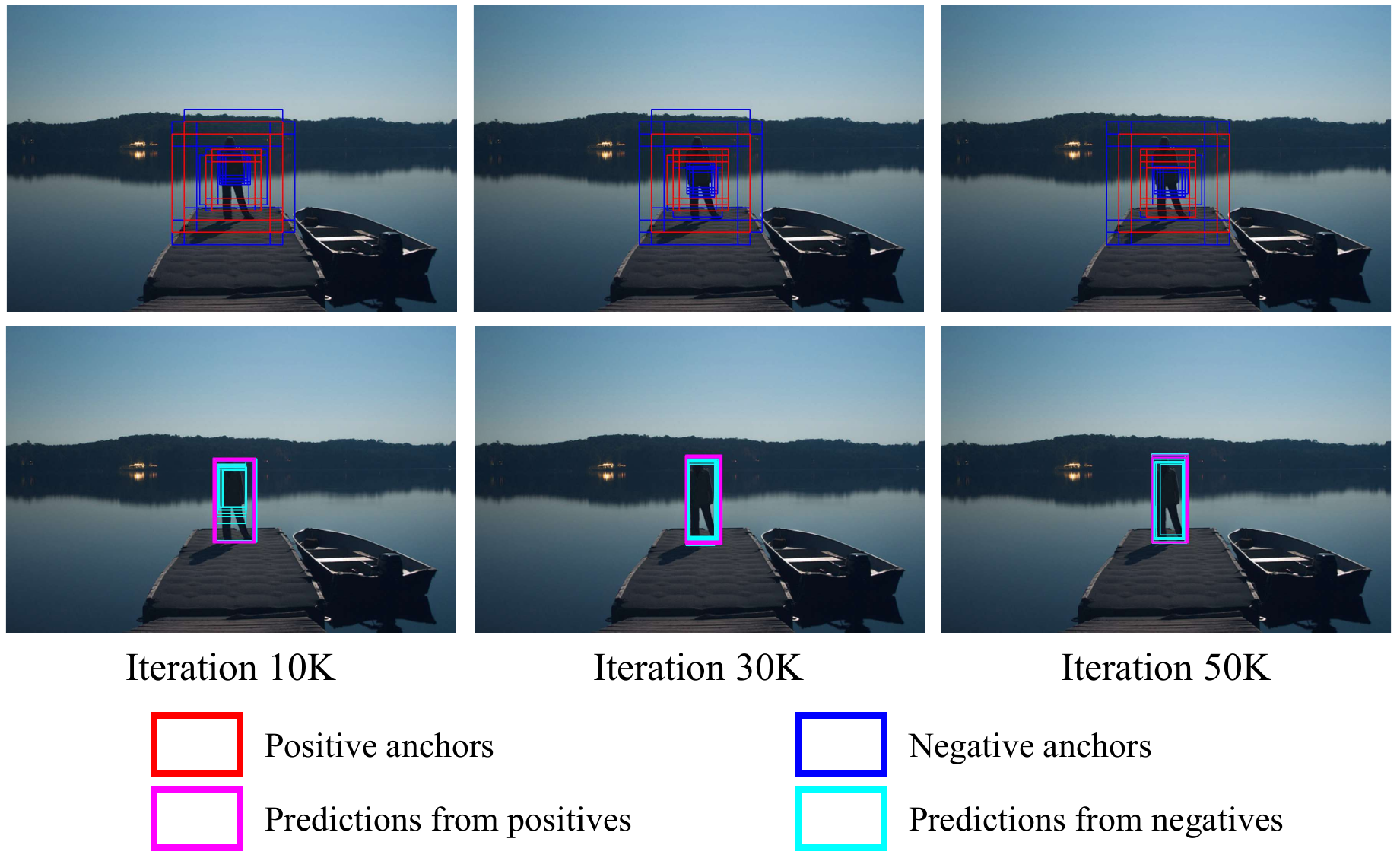}
	    \caption{Image ID: 10388}
	    \label{fig_box_vis_person}
    \end{subfigure}
    \centering
    \begin{subfigure}{.85\textwidth}
		\includegraphics[width=.95\linewidth]{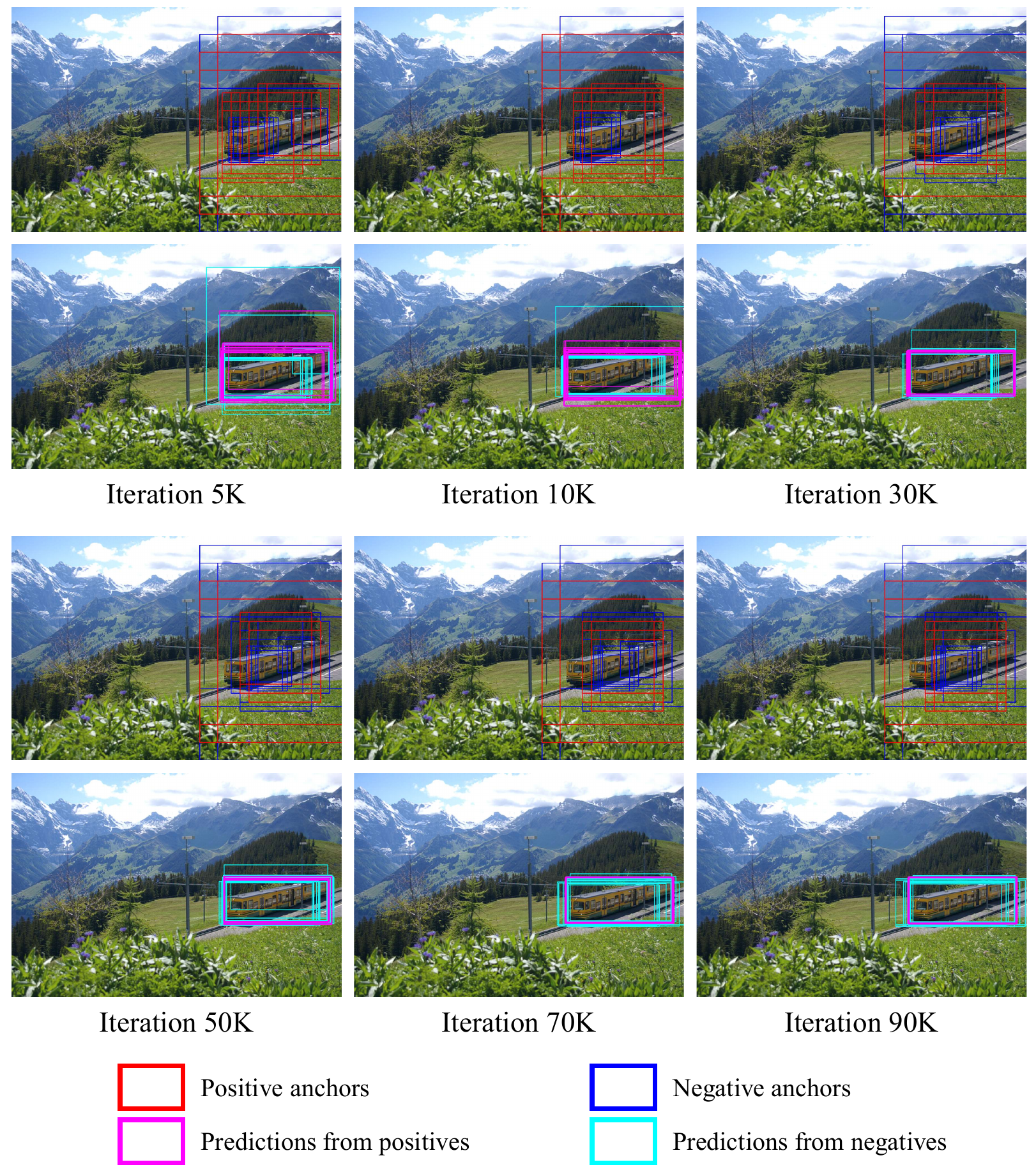}
	    \caption{Image ID: 87156}
	    \label{fig_box_vis_train}
    \end{subfigure}    
\caption{Evolution of anchor assignment and predicted boxes during training.}
\label{vis:anchors}
\end{figure}

\begin{figure}[t]
    \begin{center}
		\includegraphics[width=1.0\linewidth]{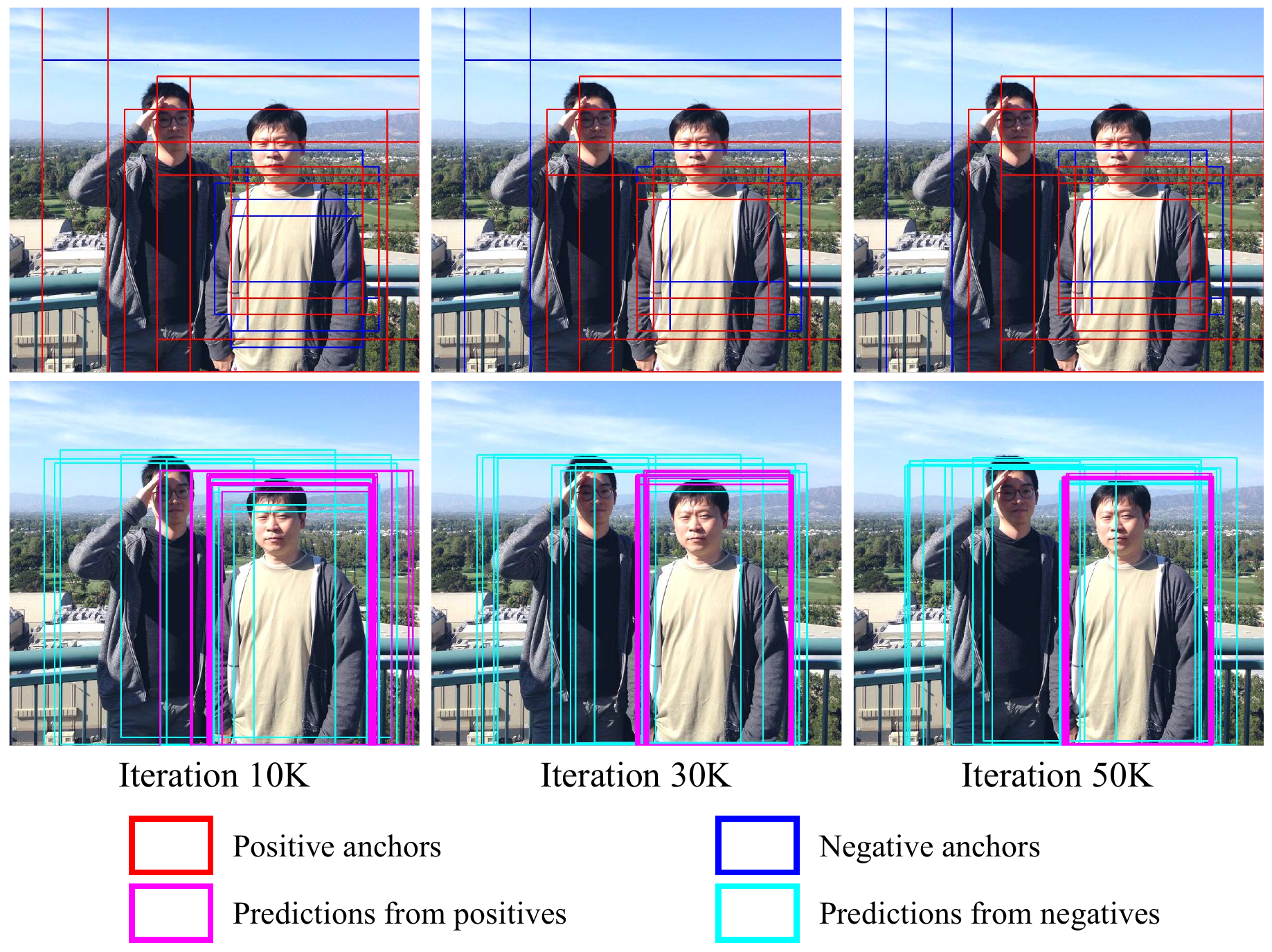}
	\end{center}
    \caption{Evolution of anchor assignment and predicted boxes on a non-COCO image.}
    \label{vis:anchors_henry}
\end{figure}

We visualize the proposed anchor assignment during training. Figure \ref{vis:anchors} shows anchor assignment results on COCO training set. Figure \ref{vis:anchors_henry} shows anchor assignment results on a non-COCO image.

\subsection{Visualization of Detection Results}
\begin{figure}[t]
    \centering
    \begin{subfigure}{.45\textwidth}
		\includegraphics[width=.85\linewidth]{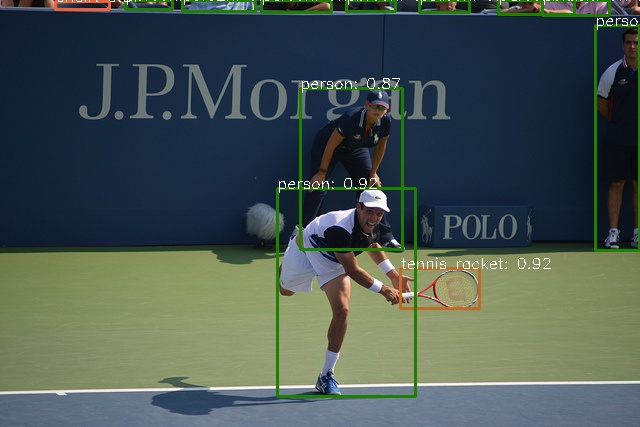}
	    \label{885}
    \end{subfigure}
    \begin{subfigure}{.45\textwidth}
		\includegraphics[width=.85\linewidth]{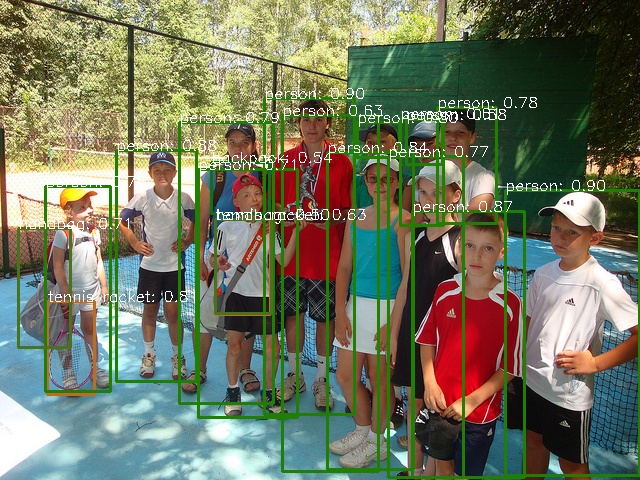}
	    \label{1000}
    \end{subfigure}
    \begin{subfigure}{.45\textwidth}
		\includegraphics[width=.85\linewidth]{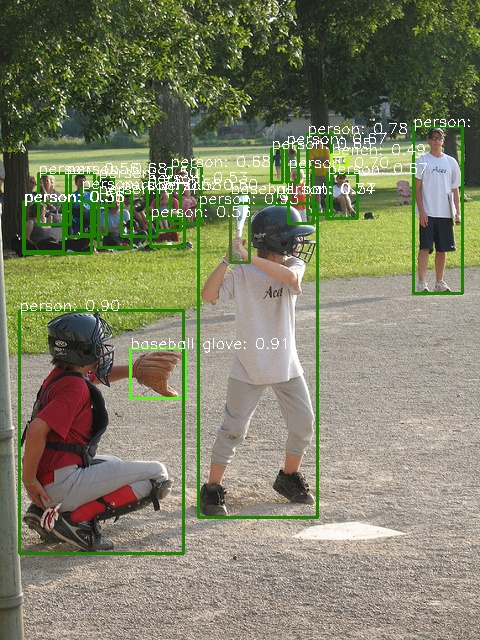}
	    \label{12639}
    \end{subfigure}
    \begin{subfigure}{.45\textwidth}
		\includegraphics[width=.85\linewidth]{fig_suppl/COCO_val2014_000000033759.jpg}
	    \label{33759}
    \end{subfigure}
    \begin{subfigure}{.45\textwidth}
		\includegraphics[width=.85\linewidth]{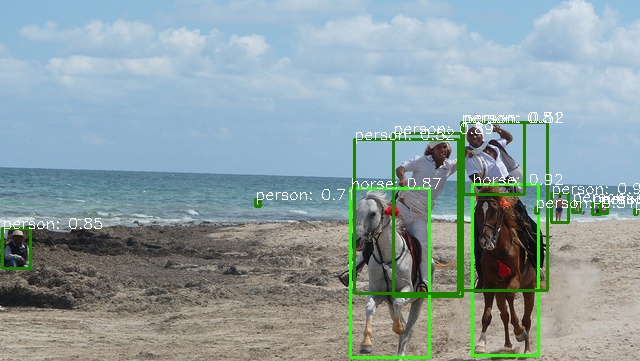}
	    \label{7281}
    \end{subfigure}
    \begin{subfigure}{.45\textwidth}
		\includegraphics[width=.85\linewidth]{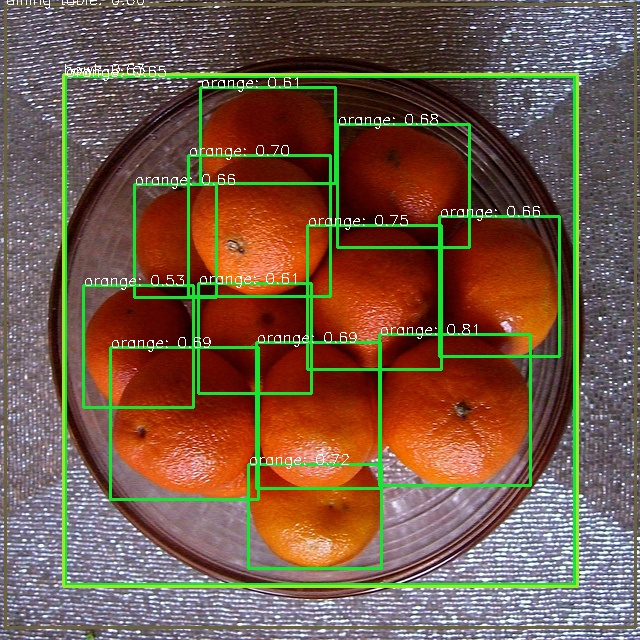}
	    \label{50896}
    \end{subfigure}
    \begin{subfigure}{.45\textwidth}
		\includegraphics[width=.85\linewidth]{fig_suppl/COCO_val2014_000000479030.jpg}
	    \label{479030}
    \end{subfigure}
    \begin{subfigure}{.45\textwidth}
		\includegraphics[width=.85\linewidth]{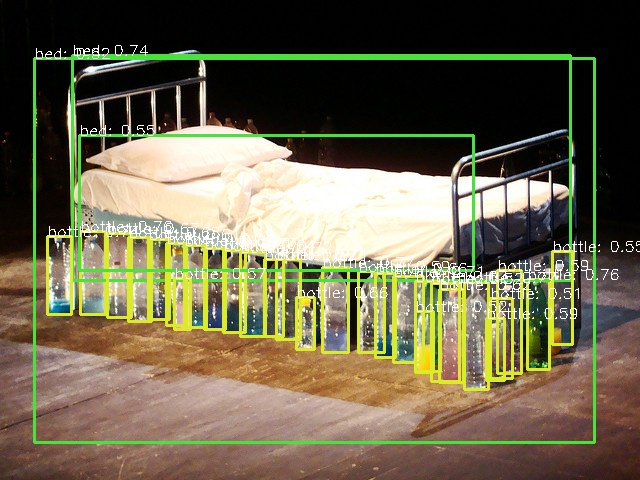}
	    \label{463842}
    \end{subfigure}
    \caption{Visualization of detection results on images of COCO minival set.}
    \label{vis:detection_results}
\end{figure}

We visualize detection results on COCO minival set in Figure \ref{vis:detection_results}.

%% file: main.bbl
\begin{thebibliography}{10}
\providecommand{\url}[1]{\texttt{#1}}
\providecommand{\urlprefix}{URL }
\providecommand{\doi}[1]{https://doi.org/#1}

\bibitem{bears}
{NPS} {P}hoto,
  \url{https://www.nps.gov/features/yell/slidefile/mammals/grizzlybear/Images/00110.jpg}

\bibitem{softnms}
Bodla, N., Singh, B., Chellappa, R., Davis, L.S.: Soft-nms--improving object
  detection with one line of code. In: Proceedings of the IEEE international
  conference on computer vision. pp. 5561--5569 (2017)

\bibitem{aploss}
Chen, K., Li, J., Lin, W., See, J., Wang, J., Duan, L., Chen, Z., He, C., Zou,
  J.: Towards accurate one-stage object detection with ap-loss. In: Proceedings
  of the IEEE Conference on Computer Vision and Pattern Recognition. pp.
  5119--5127 (2019)

\bibitem{gaussyolo}
Choi, J., Chun, D., Kim, H., Lee, H.J.: Gaussian yolov3: An accurate and fast
  object detector using localization uncertainty for autonomous driving. In:
  Proceedings of the IEEE International Conference on Computer Vision. pp.
  502--511 (2019)

\bibitem{deformableconv}
Dai, J., Qi, H., Xiong, Y., Li, Y., Zhang, G., Hu, H., Wei, Y.: Deformable
  convolutional networks. In: Proceedings of the IEEE international conference
  on computer vision. pp. 764--773 (2017)

\bibitem{objectsaspoints}
Duan, K., Bai, S., Xie, L., Qi, H., Huang, Q., Tian, Q.: Centernet: Keypoint
  triplets for object detection. In: Proceedings of the IEEE International
  Conference on Computer Vision. pp. 6569--6578 (2019)

\bibitem{fastrcnn}
Girshick, R.: Fast r-cnn. In: Proceedings of the IEEE international conference
  on computer vision. pp. 1440--1448 (2015)

\bibitem{rcnn}
Girshick, R., Donahue, J., Darrell, T., Malik, J.: Rich feature hierarchies for
  accurate object detection and semantic segmentation. In: Proceedings of the
  IEEE conference on computer vision and pattern recognition. pp. 580--587
  (2014)

\bibitem{imagenet1hour}
Goyal, P., Doll{\'a}r, P., Girshick, R., Noordhuis, P., Wesolowski, L., Kyrola,
  A., Tulloch, A., Jia, Y., He, K.: Accurate, large minibatch sgd: Training
  imagenet in 1 hour. arXiv preprint arXiv:1706.02677  (2017)

\bibitem{maskrcnn}
He, K., Gkioxari, G., Doll{\'a}r, P., Girshick, R.: Mask r-cnn. In: Proceedings
  of the IEEE international conference on computer vision. pp. 2961--2969
  (2017)

\bibitem{resnet}
He, K., Zhang, X., Ren, S., Sun, J.: Deep residual learning for image
  recognition. In: Proceedings of the IEEE conference on computer vision and
  pattern recognition. pp. 770--778 (2016)

\bibitem{varvoting}
He, Y., Zhu, C., Wang, J., Savvides, M., Zhang, X.: Bounding box regression
  with uncertainty for accurate object detection. In: Proceedings of the IEEE
  Conference on Computer Vision and Pattern Recognition. pp. 2888--2897 (2019)

\bibitem{learningnms}
Hosang, J., Benenson, R., Schiele, B.: Learning non-maximum suppression. In:
  Proceedings of the IEEE conference on computer vision and pattern
  recognition. pp. 4507--4515 (2017)

\bibitem{batchnorm}
Ioffe, S., Szegedy, C.: Batch normalization: Accelerating deep network training
  by reducing internal covariate shift. arXiv preprint arXiv:1502.03167  (2015)

\bibitem{iounet}
Jiang, B., Luo, R., Mao, J., Xiao, T., Jiang, Y.: Acquisition of localization
  confidence for accurate object detection. In: Proceedings of the European
  Conference on Computer Vision (ECCV). pp. 784--799 (2018)

\bibitem{mal}
Ke, W., Zhang, T., Huang, Z., Ye, Q., Liu, J., Huang, D.: Multiple anchor
  learning for visual object detection. arXiv preprint arXiv:1912.02252  (2019)

\bibitem{alexnet}
Krizhevsky, A., Sutskever, I., Hinton, G.E.: Imagenet classification with deep
  convolutional neural networks. In: Advances in neural information processing
  systems. pp. 1097--1105 (2012)

\bibitem{cornernet}
Law, H., Deng, J.: Cornernet: Detecting objects as paired keypoints. In:
  Proceedings of the European Conference on Computer Vision (ECCV). pp.
  734--750 (2018)

\bibitem{lecun}
LeCun, Y., Boser, B., Denker, J.S., Henderson, D., Howard, R.E., Hubbard, W.,
  Jackel, L.D.: Backpropagation applied to handwritten zip code recognition.
  Neural computation  \textbf{1}(4),  541--551 (1989)

\bibitem{noisy}
Li, H., Wu, Z., Zhu, C., Xiong, C., Socher, R., Davis, L.S.: Learning from
  noisy anchors for one-stage object detection. arXiv preprint arXiv:1912.05086
   (2019)

\bibitem{fpn}
Lin, T.Y., Doll{\'a}r, P., Girshick, R., He, K., Hariharan, B., Belongie, S.:
  Feature pyramid networks for object detection. In: Proceedings of the IEEE
  conference on computer vision and pattern recognition. pp. 2117--2125 (2017)

\bibitem{focal}
Lin, T.Y., Goyal, P., Girshick, R., He, K., Doll{\'a}r, P.: Focal loss for
  dense object detection. In: Proceedings of the IEEE international conference
  on computer vision. pp. 2980--2988 (2017)

\bibitem{coco}
Lin, T.Y., Maire, M., Belongie, S., Hays, J., Perona, P., Ramanan, D.,
  Doll{\'a}r, P., Zitnick, C.L.: Microsoft coco: Common objects in context. In:
  European conference on computer vision. pp. 740--755. Springer (2014)

\bibitem{ssd}
Liu, W., Anguelov, D., Erhan, D., Szegedy, C., Reed, S., Fu, C.Y., Berg, A.C.:
  Ssd: Single shot multibox detector. In: European conference on computer
  vision. pp. 21--37. Springer (2016)

\bibitem{yolo}
Redmon, J., Divvala, S., Girshick, R., Farhadi, A.: You only look once:
  Unified, real-time object detection. In: Proceedings of the IEEE conference
  on computer vision and pattern recognition. pp. 779--788 (2016)

\bibitem{yolo2}
Redmon, J., Farhadi, A.: Yolo9000: better, faster, stronger. In: Proceedings of
  the IEEE conference on computer vision and pattern recognition. pp.
  7263--7271 (2017)

\bibitem{yolov3}
Redmon, J., Farhadi, A.: Yolov3: An incremental improvement. arXiv preprint
  arXiv:1804.02767  (2018)

\bibitem{fasterrcnn}
Ren, S., He, K., Girshick, R., Sun, J.: Faster r-cnn: Towards real-time object
  detection with region proposal networks. In: Advances in neural information
  processing systems. pp. 91--99 (2015)

\bibitem{giou}
Rezatofighi, H., Tsoi, N., Gwak, J., Sadeghian, A., Reid, I., Savarese, S.:
  Generalized intersection over union: A metric and a loss for bounding box
  regression. In: Proceedings of the IEEE Conference on Computer Vision and
  Pattern Recognition. pp. 658--666 (2019)

\bibitem{imagenet}
Russakovsky, O., Deng, J., Su, H., Krause, J., Satheesh, S., Ma, S., Huang, Z.,
  Karpathy, A., Khosla, A., Bernstein, M., et~al.: Imagenet large scale visual
  recognition challenge. International journal of computer vision
  \textbf{115}(3),  211--252 (2015)

\bibitem{vgg}
Simonyan, K., Zisserman, A.: Very deep convolutional networks for large-scale
  image recognition. arXiv preprint arXiv:1409.1556  (2014)

\bibitem{googlenet}
Szegedy, C., Liu, W., Jia, Y., Sermanet, P., Reed, S., Anguelov, D., Erhan, D.,
  Vanhoucke, V., Rabinovich, A.: Going deeper with convolutions. In:
  Proceedings of the IEEE conference on computer vision and pattern
  recognition. pp.~1--9 (2015)

\bibitem{fcos}
Tian, Z., Shen, C., Chen, H., He, T.: Fcos: Fully convolutional one-stage
  object detection. In: Proceedings of the IEEE International Conference on
  Computer Vision. pp. 9627--9636 (2019)

\bibitem{guidedanchor}
Wang, J., Chen, K., Yang, S., Loy, C.C., Lin, D.: Region proposal by guided
  anchoring. In: Proceedings of the IEEE Conference on Computer Vision and
  Pattern Recognition. pp. 2965--2974 (2019)

\bibitem{groupnorm}
Wu, Y., He, K.: Group normalization. In: Proceedings of the European Conference
  on Computer Vision (ECCV). pp. 3--19 (2018)

\bibitem{resnext}
Xie, S., Girshick, R., Doll{\'a}r, P., Tu, Z., He, K.: Aggregated residual
  transformations for deep neural networks. In: Proceedings of the IEEE
  conference on computer vision and pattern recognition. pp. 1492--1500 (2017)

\bibitem{metaanchor}
Yang, T., Zhang, X., Li, Z., Zhang, W., Sun, J.: Metaanchor: Learning to detect
  objects with customized anchors. In: Advances in Neural Information
  Processing Systems. pp. 320--330 (2018)

\bibitem{reppoints}
Yang, Z., Liu, S., Hu, H., Wang, L., Lin, S.: Reppoints: Point set
  representation for object detection. In: Proceedings of the IEEE
  International Conference on Computer Vision. pp. 9657--9666 (2019)

\bibitem{iouloss}
Yu, J., Jiang, Y., Wang, Z., Cao, Z., Huang, T.: Unitbox: An advanced object
  detection network. In: Proceedings of the 24th ACM international conference
  on Multimedia. pp. 516--520 (2016)

\bibitem{atss}
Zhang, S., Chi, C., Yao, Y., Lei, Z., Li, S.Z.: Bridging the gap between
  anchor-based and anchor-free detection via adaptive training sample
  selection. arXiv preprint arXiv:1912.02424  (2019)

\bibitem{refinedet}
Zhang, S., Wen, L., Bian, X., Lei, Z., Li, S.Z.: Single-shot refinement neural
  network for object detection. In: Proceedings of the IEEE conference on
  computer vision and pattern recognition. pp. 4203--4212 (2018)

\bibitem{freeanchor}
Zhang, X., Wan, F., Liu, C., Ji, R., Ye, Q.: Freeanchor: Learning to match
  anchors for visual object detection. In: Advances in Neural Information
  Processing Systems. pp. 147--155 (2019)

\bibitem{m2det}
Zhao, Q., Sheng, T., Wang, Y., Tang, Z., Chen, Y., Cai, L., Ling, H.: M2det: A
  single-shot object detector based on multi-level feature pyramid network. In:
  Proceedings of the AAAI Conference on Artificial Intelligence. vol.~33, pp.
  9259--9266 (2019)

\bibitem{extremepoints}
Zhou, X., Zhuo, J., Krahenbuhl, P.: Bottom-up object detection by grouping
  extreme and center points. In: Proceedings of the IEEE Conference on Computer
  Vision and Pattern Recognition. pp. 850--859 (2019)

\bibitem{fsaf}
Zhu, C., He, Y., Savvides, M.: Feature selective anchor-free module for
  single-shot object detection. In: Proceedings of the IEEE Conference on
  Computer Vision and Pattern Recognition. pp. 840--849 (2019)

\end{thebibliography}
